# Question answering system of bridge design specification based on large language model

Leye Zhang[1], Xiangxiang Tian[1], Hongjun Zhang[2]

(1. Jiangsu College of Finance & Accounting, Lianyungang, 222061, China;

2. Wanshi Antecedence Digital Intelligence Traffic Technology Co., Ltd, Nanjing, 210016, China)

**Abstract:** This paper constructs question answering system for bridge design specification based on large language model. Three implementation schemes are tried: full fine-tuning of the BERT pretrained model, parameter-efficient fine-tuning of the BERT pretrained model, and self-built language model from scratch. Through the self-built question and answer task dataset, based on the TensorFlow and Keras deep learning platform framework, the model is constructed and trained to predict the start position and end position of the answer in the bridge design specification given by the user. The experimental results show that full fine-tuning of the BERT pretrained model achieves 100% accuracy in the training-dataset, validation-dataset and test-dataset, and the system can extract the answers from the bridge design specification given by the user to answer various questions of the user; While parameter-efficient fine-tuning of the BERT pretrained model and self-built language model from scratch perform well in the training-dataset, their generalization ability in the test-dataset needs to be improved. The research of this paper provides a useful reference for the development of question answering system in professional field.

## 0 Introduction

At work, professionals often need to query information in the literature. The most common keyword search technology has limitations such as information loss, too much returned information, and irrelevant information [1]. The traditional question answering system (QA) based on structured data or frequently asked Questions(FAQ) relies on rules, templates or a limited set of built-in question answers, which is difficult to cope with complex and changeable user needs and has high development and maintenance costs. In recent years, the rapid development of large language model (LLM) has brought new life to the question answering system based on natural language processing (NLP), which is expected to become a new generation of search engine.

ChatGPT and other generative pre-trained models have the ability of language understanding and text generation, and can answer users' questions, but the generated results will have hallucinations and deviations, which need further fine-tuning before they can be used in the professional field. At present, the research of question answering system based on large language model has been actively carried out in many professional fields. Based on large language model, zhangjinying built an intelligent question answering system for electric power knowledge [2]; Zhangchunhong studied educational question answering system based on large language model [3]; Lei Tianfeng studied question and answer system for competitive vehicle configuration based on large language model [4]; Based on large language model, Wang Ting constructed question and answer system for fruit and vegetable agricultural technology knowledge [5]; Dingzhikun designed BIM forward design question answering system based on large language model [6]; MM Lucas proposed a new prompt method to improve inference ability of medical question answering system based on large language model [7]; Wonjin Yoon studied biomedical question answering system based on large language model [8]. However, the application of large language model to question answering system in bridge design has not been reported.

Self built question and answer task dataset in this paper. Question and answer system of bridge design specification is implemented by three schemes: full fine-tuning of the BERT pretrained model, parameter-efficient fine-tuning of the BERT pretrained model, and self-built language model from scratch. The model can extract answers from the bridge design specification given by the user in advance to answer various questions of the user (open source address of this article's dataset and source code: https://github.com/ zhangleye/Bridge-LLM-QA).

# 1 Introduction to large language model and question answering system
## 1.1 Large language model

1. Large language model is based on Transformer and uses large number of text data training to understand and generate natural language text [9]. It shows amazing versatility and can handle a variety of natural language tasks, such as text classification, question and answer, dialogue and so on. It is regarded as a possible way to achieve Artificial General Intelligence (AGI). However, the current technology level is still far from the real AGI.

At present, AI is characterized by fragmentation and diversification in terms of industry, business scenarios and demands, and generally adopts the production mode of "pre-trained large model + downstream task fine-tuning". After pre-trained (self-supervised learning), large model obtains the general ability to complete various tasks; Then fine-tuning (supervised learning) to adapt to specific tasks. This is somewhat similar to personnel cultivating program in the current human university, which is to study public basic courses in the early stage and professional skills courses in the later stage.

2. Restricted by the hardware performance, this paper uses a small-scale "bert-base-chinese" pre-trained model, which is configured with L=12 (the number of Transformer encoders in series), A=12 (the number of Transformer encoder channels), H=768 (the dimension of word embedding space), and the total number of network parameters is 110 million [10]. The vocabulary has more than 7000 Chinese characters, which meets the requirements of this study.

The BERT model is pre-trained for two specific tasks on a huge corpus. The two tasks are masked language and next sentence prediction. When the BERT model is used for actual tasks, only the output layer of a specific task is added, and then fine-tuning can be done. It is very simple.

3. the common loading methods of pre-trained large model are: loading through PyTorch, TensorFlow and the Transformers code of Hugging Face.

## 1.2 Question answering system

Question answering system is an advanced form of information retrieval system. The main reason for the rise of its research is people's demand for fast and accurate access to information. Question answering system is a research direction in the field of artificial intelligence and natural language processing, which has attracted much attention and broad development prospects. At present, the representative technical routes are: question answering system based on structured data, question answering system based on frequently asked questions, and question answering system based on natural language processing [11].

The question answering system based on natural language processing can directly generate text from the language model to answer (open-ended, there may be no standard answer), and can also find answers from the Internet or reading materials (abstract or extraction). For the professional field, under the current technical level, the extraction type is more suitable. It extracts text fragments from the given documents to answer (just as the examinee is doing a Chinese or English reading comprehension question, the examinee understands the given reading material, and answers the question according to the specific details of the reading material).

# 2 Software and hardware environments, and self-built question and answer task dataset
## 2.1 Software and hardware environments

The operating system is Windows 11 (for TensorFlow 2.10, Keras 2.10) and WSL Ubuntu 22.04 (for

TensorFlow 2.16.1, Keras 3.4.0, Keras_nlp 0.12.1), CUDA version 12.3. The CPU is AMD ryzen 52600, the memory is 64g DDR4, and the graphics card is RTX 4060 Ti 16G. Cloud computing is not used.

## 2.2 Self-built question and answer task dataset

At present, there are more than ten domestic current specifications for the design of highway and municipal bridges. Restricted by the cost, this paper only takes the text content of Chapter 3 of 《General Specifications for Design of Highway Bridges and Culverts》（JTG D60-2015）[12] to make a question and answer task dataset.

The dataset is in the xlsx file format, with one sample per row and a total of 226 samples. The header is: context, question, answer, index_of_answer. Context is the given specification text content (reading comprehension materials), question is the question, answer is the standard answer (the text segment extracted from the context), and index_of_answer is the index number of the standard answer (because the text segment of the standard answer will appear many times in the context, but only one part meets the requirements, so use this parameter to locate). The dataset is as follows: (Only 3 samples are shown and translate Chinese into English)

Tab.1 Dataset of question answering

| | A | B | C | D |
|---|---|---|---|---|
| 1 | context | question | answer | index_of_answer |
| 2 | 3.1.1 Highway bridges and culverts shall be designed as a whole according to highway functions and technical grades, considering factors such as local conditions, local materials, convenience for construction and maintenance, and shall meet the requirements of normal traffic load within the design service life.<br>3.1.2 The alignment design of highway bridges and culverts shall meet the following requirements:<br>1 The alignment design of small and medium-sized bridges and culverts shall meet the overall requirements of route design.<br>2 The alignment design of super large and large bridges shall comprehensively consider the overall trend of the route, geology, terrain, sensitive traffic, navigation, existing building facilities, environmentally sensitive areas and other factors.<br>3 Super large and large bridges should adopt higher index of horizontal curve, and the vertical section should not be designed as flat slope or concave curve.<br>3.1 3 Structure of highway bridges and culverts shall be designed according to the ultimate limit state and serviceability limit state. | What factors should be considered in the overall design of highway Bridges and culverts? | such as local conditions, local materials, convenience for construction and maintenance | 0 |
| 3 | 3.4.2 The layout of pavement, bicycle lane and barrier facilities on the bridge deck shall comply with the following provisions:<br>1 Sidewalks should not be set for bridges on expressways. The setting of sidewalks and bicycle lanes on bridges on class I, II, III and IV highways shall be determined as required, and shall be coordinated with the layout of the front and rear routes. Guardrails or curbs and other separation facilities shall be provided between sidewalks, bicycle lanes and carriageways. The width of a bicycle lane shall be 1.0m; When a bicycle lane is set separately, it should not be less than the width of two bicycle lanes. The width of sidewalk should be 1.0m; When it is greater than 1.0m, it will be increased by 0.5m differential. Sidewalks may not be provided for submersible bridges and overflow Pavements.<br>2 The width of the slow lane mainly used by tractors or animal powered vehicles shall be determined according to the local tractor or animal powered vehicle type and traffic volume; When set along one side of the bridge, it shall not be less than the width required for two-way traffic.<br>3 The setting of bridge guardrails shall comply with the relevant provisions of 《Design Specifications for Highway Safety Facilities》（JTG D81）.<br>4 The height of curbs can be 0.25~0.35m. When crossing rapids, broad rivers, deep valleys, important roads, railways, and major shipping lanes, or when the bridge decks are often covered with snow and ice, the height of the curbs should be taken as a larger value. | How wide should the sidewalk on the bridge deck be? | 1.0m | 1 |
| 4 | 3.4 4 Culverts should be designed as non pressure type. The clear height from the apex of the pressureless culvert to the standard water level of the design flood frequency in the culvert shall comply with the provisions of table 3.4.4.<br>3.4.5 The clearance under the overpass bridges shall meet the following requirements:<br>1 In addition to complying with the provisions of article 3.4.1 bridges and culverts clearance in this specification, the clearance and hole arrangement under the overpass bridge of highway to highway shall also meet the requirements of sight distance and front information identification of the highway under the bridge, and its structural form shall be coordinated with the surrounding environment.<br>2 When the railway crosses the highway, the clearance and hole arrangement under the overpass bridge shall meet the requirements of sight distance and front information identification of the highway under the bridge in addition to the provisions of bridge and culvert clearance in article 3.4.1 of this specification.<br>3. The clearance under the overpass bridge of rural roads and highways is:<br>1) When a rural road crosses over a highway, the clearance under the overpass bridge shall comply with article 3 4.1 provisions on construction clearance;<br>2) When a rural road passes under a highway, its clearance can be determined according to the local traffic and intersection conditions. The clear height of the pedestrian passage should be greater than or equal to 2.2M, and the clear width should be greater than or equal to 4.0m;<br>3) The clear height of the passage for animal powered vehicles and tractors shall be greater than or equal to 2.7m, and the clear width shall be greater than or equal to 4.0m; | What is the clear width of the passage for animal powered vehicles and tractors under the overpass bridge greater than or equal to? | 4.0m | 1 |

Because the BERT pre-trained model used in this paper has a maximum input length of 512 tokens (including punctuation marks, parentheses, decimal points, etc.), and 509 available tokens for users (deducting [CLS] and [SEP] marks), it is necessary to set a context of no more than 474 tokens and a question of no more than 35 tokens.

## 3 Implementation scheme 1 of question answering system for bridge design specification: full fine-tuning of the BERT pretrained model

### 3.1 Download vocabulary and create tokenizer

Download the vocabulary of "bert-base-chinese" from the Transformers library of Hugging Face, and then create tokenizer. A Chinese character is a token.

### 3.2 Loading dataset and data preprocessing

The read_excel() function of pandas module is used to read the dataset. The dataset was randomly divided into training-dataset (200 samples), validation-dataset (20 samples) and test-dataset (6 samples).

Then convert the text format to the integer format vector specified by the BERT model. Each sample consists of sample input X and label y. The sample input x consists of "input_ids", "token_type_ids" and "attention_mask". Label y consists of "start_token_idx" and "end_token_idx".

Where: "input_ids" is composed of the context and question of the sample; "token_type_ids" is a segmented mark, the context is taken as 0, the question is taken as 1, and the tail filling part is taken as 0; The filled part of the tail of "attention_mask" is taken as 0, and the others are taken as 1; "start_token_idx" is the index number of the starting position of the sample answer in the context; "end_token_idx" is the index number of the end position of the sample answer in the context.

## 3.3 Model Construction

In question and answer task, given a context and a question, the model needs to find the answer related to the question from the context. This requires adding a header model (output layer) for question and answer task on the basis of the BERT model. The header model contains two parallel Dense layers (softmax activation), which are used to predict the start position and end position of the answer to the output of the BERT model.

Here, download "bert-base-chinese" from the Transformers library of Hugging Face. (If the question answering system model based on BERT is downloaded from the Transformers library, the original model already contains a header model, you can directly fine-tuning it at this time.)

Build the model based on Python 3.10 programming language, TensorFlow 2.10 and Keras 2.10 deep learning platform framework. See the following figure for the code:

```python
def create_model():
    encoder = TFBertModel.from_pretrained("bert-base-chinese")
    input_ids = layers.Input(shape=(max_len,), dtype=tf.int32)
    token_type_ids = layers.Input(shape=(max_len,), dtype=tf.int32)
    attention_mask = layers.Input(shape=(max_len,), dtype=tf.int32)
    embedding = encoder(input_ids, token_type_ids=token_type_ids, attention_mask=attention_mask)[0]
    start_logits = layers.Dense(1, name="start_logit", use_bias=False)(embedding)
    start_logits = layers.Flatten()(start_logits)
    end_logits = layers.Dense(1, name="end_logit", use_bias=False)(embedding)
    end_logits = layers.Flatten()(end_logits)
    start_probs = layers.Activation(keras.activations.softmax)(start_logits)
    end_probs = layers.Activation(keras.activations.softmax)(end_logits)
    model = keras.Model( inputs=[input_ids, token_type_ids, attention_mask], outputs=[start_probs, end_probs] )
    loss = keras.losses.SparseCategoricalCrossentropy(from_logits=False)
    optimizer = keras.optimizers.Adam(lr=learning_rate)
    model.compile(optimizer=optimizer, loss=[loss, loss], metrics=['accuracy'])
    return model
```

Fig.1 Code for model of scheme one

The model summary is shown in the following table:

Tab.2 Model summary of scheme one

| Layer (type) | Output Shape | Param # | Connected to |
|---|---|---|---|
| input_1 (InputLayer) | [(None, 512)] | 0 | [] |
| input_3 (InputLayer) | [(None, 512)] | 0 | [] |
| input_2 (InputLayer) | [(None, 512)] | 0 | [] |
| tf_bert_model (TFBertModel) | TFBaseModelOutputWithPoolingAndCrossAttentions( last_hidden_state=(None, 512,768), pooler_output=(None, 768), past_key_values=None, hidden_states=None, attentions=None, cross_attentions=None) | 102267648 | ['input_1[0][0]', 'input_3[0][0]', 'input_2[0][0]'] |
| start_logit (Dense) | (None, 512, 1) | 768 | ['tf_bert_model[0][0]'] |
| end_logit (Dense) | (None, 512, 1) | 768 | ['tf_bert_model[0][0]'] |
| flatten (Flatten) | (None, 512) | 0 | ['start_logit[0][0]'] |
| flatten_1 (Flatten) | (None, 512) | 0 | ['end_logit[0][0]'] |
| activation (Activation) | (None, 512) | 0 | ['flatten[0][0]'] |
| activation_1 (Activation) | (None, 512) | 0 | ['flatten_1[0][0]'] |
| Total params: 102,269,184 | | | |
| Trainable params: 102,269,184 | | | |
| Non-trainable params: 0 | | | |

The loss function uses "SparseCategoricalCrossentropy"。

### 3.4 Fine-tuning and testing

The model receives context and question as input, and the model output is the prediction of the start and end positions of the answer. By minimizing the difference between the prediction results and label y, the weight parameters of the BERT basic model and header model are optimized. The "Adam" optimizer is used to update the parameters of the neural network, and the monitoring index is "accuracy". Use the callback function to save the weight of each round of model.

See the following figure for fine-tuning loss and accuracy (the first 28 epochs):

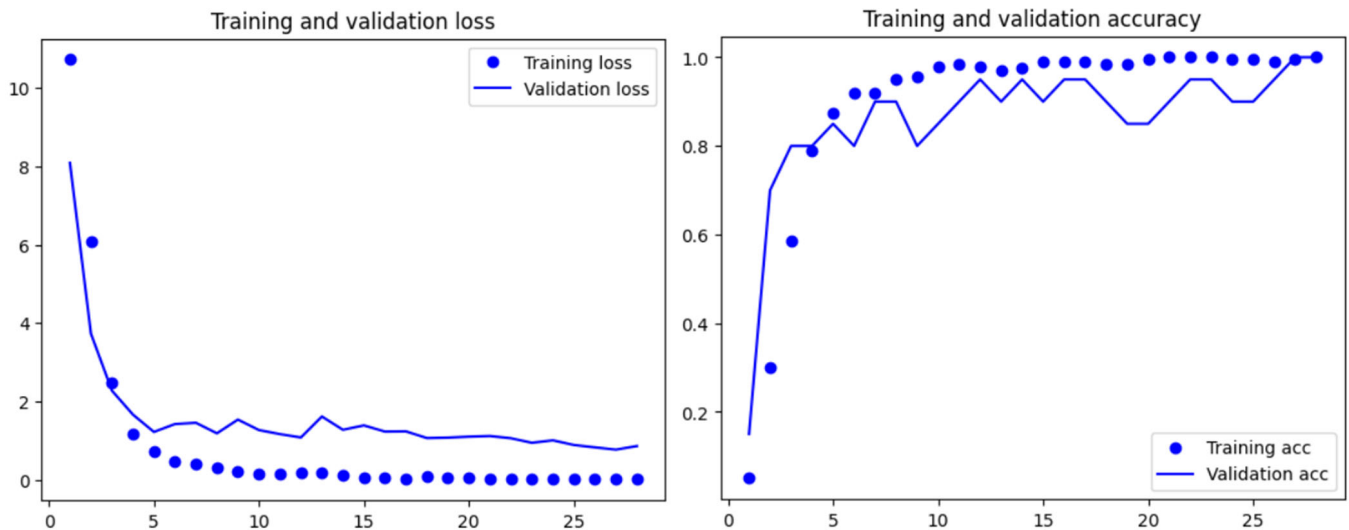

Fig.2 Fine-tuning loss and accuracy curves of scheme one

After 28 epochs of fine-tuning, the accuracy of the model in training-dataset, validation-dataset and test-dataset is 100%. It can be seen that the performance of this scheme is very good.

After testing, the basic model of BERT is frozen before fine-tuning, and only the header model is fine-tuning, which has poor performance.

## 4 Implementation scheme 2 of question answering system for bridge design specification : parameter-efficient fine-tuning of the BERT pretrained model

Unlike full fine-tuning, parameter-efficient fine-tuning only trains a small part of the parameters of the model. Lora fine-tuning is used here. Lora's idea is very simple, that is, add a bypass next to the original pretrained model to perform the operation of dimension reduction and dimension upgrading.

### 4.1 Download vocabulary and create tokenizer
Same as scheme 1.

### 4.2 Loading dataset and data preprocessing
Same as scheme 1.

### 4.3 Model Construction
Here, download "bert-base-chinese" from the Keras_nlp library, and call the enable_lora() method to freeze all the weights of the BERT basic model and establish the Lora bypass. The header model is the same as the scheme 1.

Build the model based on Python 3.11 programming language, TensorFlow 2.16.1 and Keras 3.4 deep learning platform framework. See the following figure for the code:

```python
def create_model():
    encoder = keras_nlp.models.Backbone.from_preset("bert_base_zh", load_weights=True)
    encoder.enable_lora(rank=16) #rank=4、8、16
    input_ids = layers.Input(shape=(max_len,), dtype=tf.int32)
    token_type_ids = layers.Input(shape=(max_len,), dtype=tf.int32)
    attention_mask = layers.Input(shape=(max_len,), dtype=tf.int32)
    embedding = encoder( {"token_ids":input_ids, "segment_ids":token_type_ids, "padding_mask":attention_mask} )
    start_logits = layers.Dense(1, name="start_logit", use_bias=False)(embedding['sequence_output'])
    start_logits = layers.Flatten()(start_logits)
    end_logits = layers.Dense(1, name="end_logit", use_bias=False)(embedding['sequence_output'])
    end_logits = layers.Flatten()(end_logits)
    start_probs = layers.Activation(keras.activations.softmax)(start_logits)
    end_probs = layers.Activation(keras.activations.softmax)(end_logits)
    model = keras.Model( inputs=[input_ids, token_type_ids, attention_mask], outputs=[start_probs, end_probs] )
    loss = keras.losses.SparseCategoricalCrossentropy(from_logits=False)
    optimizer = keras.optimizers.Adam(learning_rate=learning_rate)
    model.compile(optimizer=optimizer, loss=[loss, loss],metrics=['accuracy','accuracy'] )
    return model
```

Fig.3 Code for model of scheme two

The model summary is shown in the following table:

Tab.3 Model summary of scheme two

| Layer (type) | Output Shape | Param # | Connected to |
| --- | --- | --- | --- |
| input_layer_2 (InputLayer) | (None, 512) | 0 | - |
| input_layer_1 (InputLayer) | (None, 512) | 0 | - |
| input_layer (InputLayer) | (None, 512) | 0 | - |
| bert_backbone (BertBackbone) | [(None, 768), (None, 512, 768)] | 105,831,1… | input_layer_2[0]… input_layer_1[0]… input_layer[0][0] |
| start_logit (Dense) | (None, 512, 1) | 768 | bert_backbone[0]… |
| end_logit (Dense) | (None, 512, 1) | 768 | bert_backbone[0]… |
| flatten (Flatten) | (None, 512) | 0 | start_logit[0][0] |
| flatten_1 (Flatten) | (None, 512) | 0 | end_logit[0][0] |
| activation (Activation) | (None, 512) | 0 | flatten[0][0] |
| activation_1 (Activation) | (None, 512) | 0 | flatten_1[0][0] |

Total params: 105,832,704 (403.72 MB)
Trainable params: 3,583,488 (13.67 MB)
Non-trainable params: 102,249,216 (390.05 MB)

The loss function uses "SparseCategoricalCrossentropy"。

## 4.4 Fine-tuning and testing

The "Adam" optimizer is used to update the neural network parameters of Lora model and header model, and the monitoring index is "accuracy". Use the callback function to save the weight of each round of model.

See the following figure for fine-tuning loss and accuracy (the first 25 epochs):

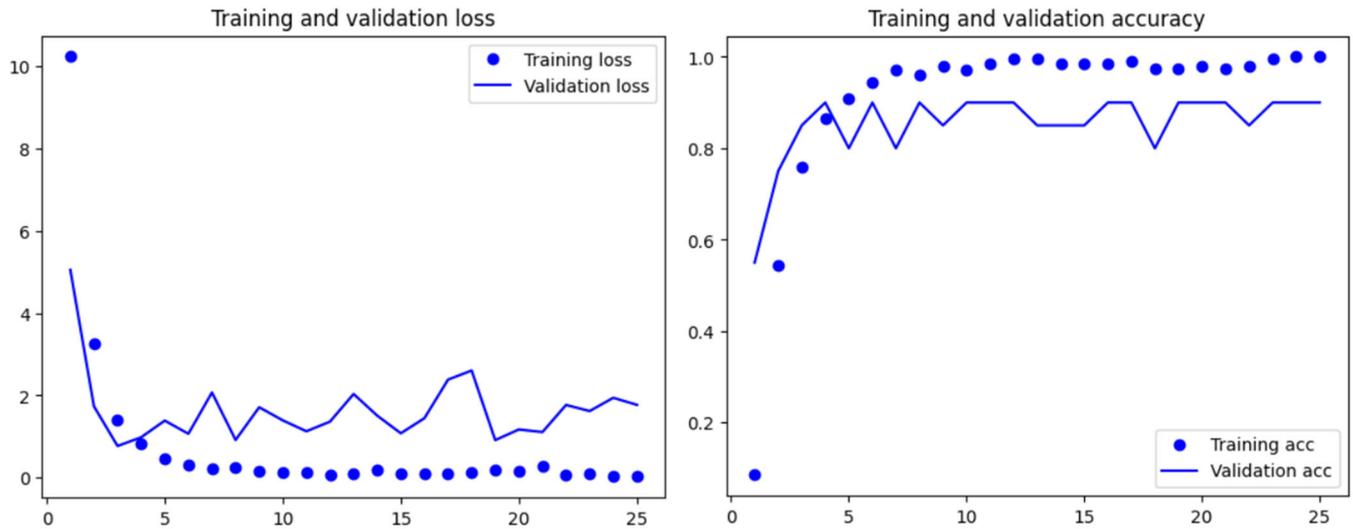

Fig.4 Fine-tuning loss and accuracy curves of scheme two

After 25 epochs of fine-tuning, the accuracy of the model in training-dataset, validation-dataset and test-dataset is 100%, 90% and 67% respectively. It can be seen that the generalization performance of this scheme is poor, which is related to the small size of the dataset and the fine-tuning method.

## 5 Implementation scheme 3 of question answering system for bridge design specification: self-built language model from scratch

If the large model is trained directly on the final task dataset without pre-training, it will lack generalization performance. The reason is that the production cost of supervised learning dataset is high, and the number of samples is usually very small, so it can not meet the learning requirement of large models. At this time, pre-training comes in handy. It first conducts self supervised learning on a large and low-cost dataset, and then transfers the general knowledge learned to the final task.

Here, we build the Transformer language model from scratch, then use a medium-sized unlabeled Chinese News dataset for pre-training, and finally fine-tuning it on question and answer task dataset.

### 5.1 Download the pre-training dataset

Download "CNewSum_v2.jsonl" and "prose.jsonl" at "https://huggingface.co/datasets/YeungNLP/firefly-pretrain-dataset". "CNewSum_v2.jsonl" is a news dataset with 275596 samples for pre-training. "Prose.jsonl" is a prose dataset with 658 samples for validation.

### 5.2 Download vocabulary and create tokenizer

Same as scheme 1.

### 5.3 Pre-training task

The mask language task is used for pre-training, that is, the sample randomly covers 15% of the words, and the training model is used to predict the covered words. This is equivalent to a cloze task. The loss function uses "SparseCategoricalCrossentropy".

### 5.4 Loading pre-training dataset and data preprocessing

The read_json() function of pandas module is used to read the pre-training dataset. Then convert the text format to an integer format vector. Each sample consists of sample input X and label y. The sample input x consists of "token_ids" and "mask_positions". Label y consists of "mask_ids" and "mask_weights".

Where: "token_ids" is the integer sequence after the sample is masked; "Mask_positions" is the masked position; "Mask_ids" is the original token id at the mask position; "Mask_weights" is the mark, the tail filling part is taken as 0, and the others are taken as 1.

### 5.5 Transformer language model construction

The model is built based on Python 3.11 programming language, TensorFlow 2.16.1 and Keras 3.4 deep learning platform framework. The pre-training sample (one-dimensional vector) is first transformed into a sequence (two-dimensional matrix) through the embedding layer (integrating position embedding). Then three Transformer encoders are used to realize context awareness. See the following figure for code:

```python
inputs = keras.Input(shape=(SEQ_LENGTH,), dtype="int32")
embedding_layer = keras_nlp.layers.TokenAndPositionEmbedding(
    vocabulary_size=tokenizer.vocabulary_size(),
    sequence_length=SEQ_LENGTH,
    embedding_dim=MODEL_DIM, )
outputs = embedding_layer(inputs)
outputs = keras.layers.LayerNormalization(epsilon=NORM_EPSILON)(outputs)
outputs = keras.layers.Dropout(rate=DROPOUT)(outputs)
for i in range(NUM_LAYERS):
    outputs = keras_nlp.layers.TransformerEncoder(
        intermediate_dim=INTERMEDIATE_DIM,
        num_heads=NUM_HEADS,
        dropout=DROPOUT,
        layer_norm_epsilon=NORM_EPSILON,
    )(outputs)
encoder_model = keras.Model(inputs, outputs)
```

Fig.5 Code for model of scheme three

The model summary is shown in the following table:

Tab.4 Model summary of scheme three

| Layer (type) | Output Shape | Param # |
| --- | --- | --- |
| input_layer_5 (InputLayer) | (None, 512) | 0 |
| token_and_position_embedding_4 (TokenAndPositionEmbedding) | (None, 512, 512) | 11,079,680 |
| layer_normalization_4 (LayerNormalization) | (None, 512, 512) | 1,024 |
| dropout_16 (Dropout) | (None, 512, 512) | 0 |
| transformer_encoder_12 (TransformerEncoder) | (None, 512, 512) | 2,078,172 |
| transformer_encoder_13 (TransformerEncoder) | (None, 512, 512) | 2,078,172 |
| transformer_encoder_14 (TransformerEncoder) | (None, 512, 512) | 2,078,172 |

Total params: 17,315,220 (66.05 MB)
Trainable params: 17,315,220 (66.05 MB)
Non-trainable params: 0 (0.00 B)

## 5.6 Pre-Training

The "Adam" optimizer is used to update the parameters of the neural network, and the monitoring index is "accuracy".

See the following figure for pre-training loss and accuracy:

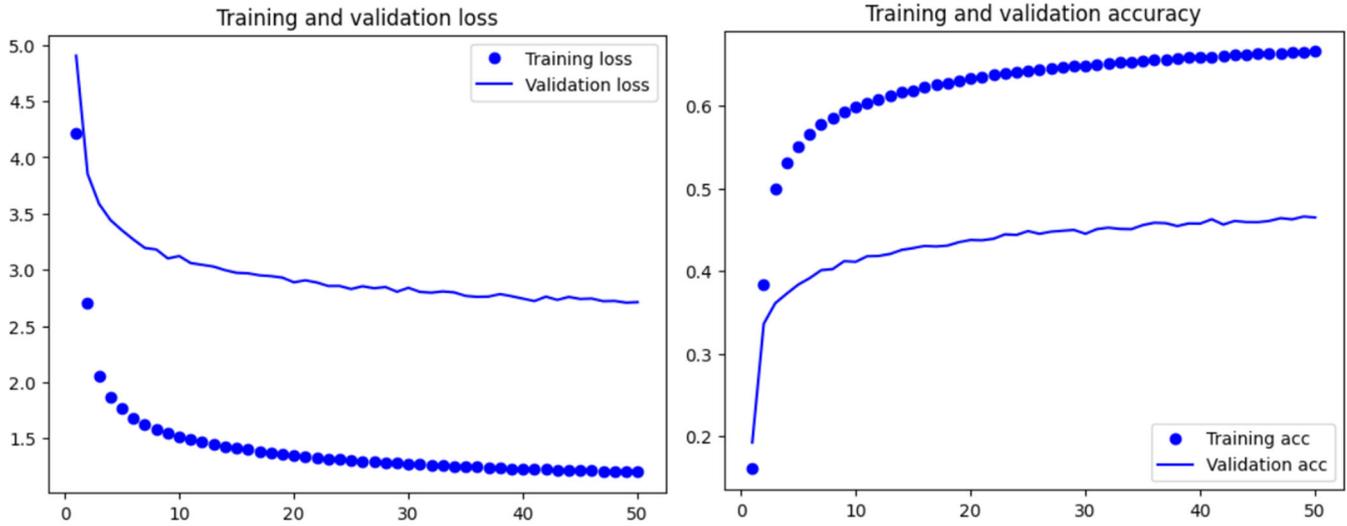
Fig.6 Pre-training loss and accuracy curves of scheme three

### 5.7 Fine-tuning for question answering task

Full fine-tuning is adopted, and the method is the same as the scheme 1.

See the following figure for fine-tuning loss and accuracy:

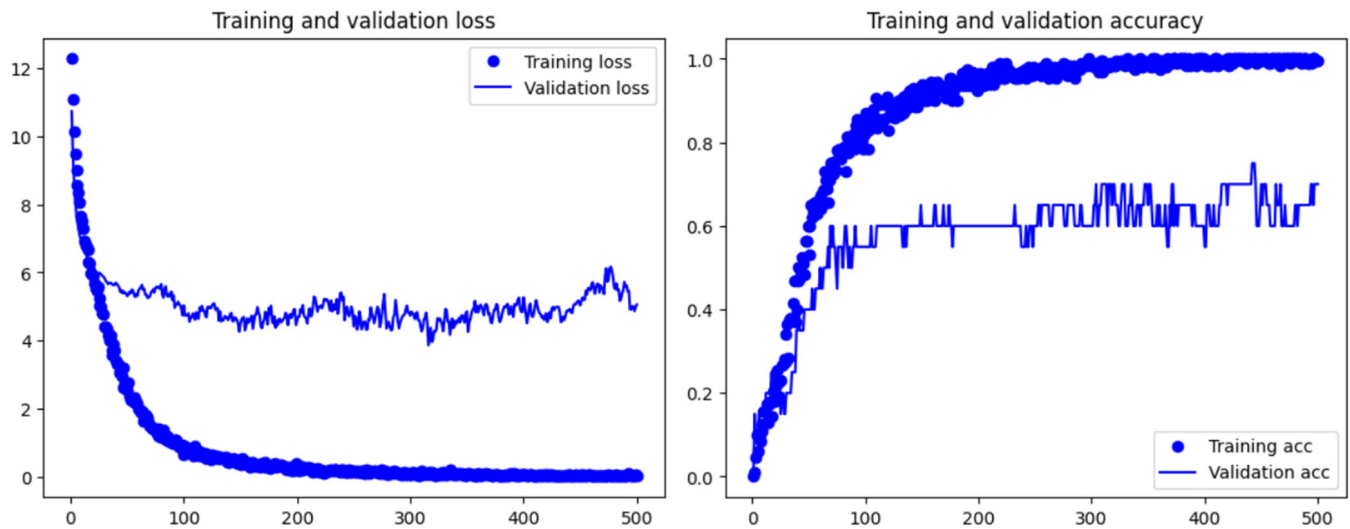
Fig.7 Fine-tuning loss and accuracy curves of scheme three

After 500 epochs of fine-tuning, the accuracy of the model in training-dataset, validation-dataset and test-dataset is 100%, 70% and 67% respectively. It can be seen that the generalization performance of this scheme is poor, which is related to the small size of the dataset and the quality of the pre-trained model.

## 6  Conclusion

In this paper, we build our own question and answer task dataset, and use three different schemes to realize bridge design specification question and answer system based on large language model. It shows the application potential of question and answer system based on large language model in professional field.

Full fine-tuning of the BERT pretrained mode achieves 100% accuracy in the training-dataset, validation-dataset and test-dataset, and has excellent performance, which verifies the applicability of large language model in the question answering system. Although parameter-efficient fine-tuning of the BERT pretrained model and self-built language model from scratch meet the challenges of generalization performance on the test-dataset, the generalization ability of the model can be improved by expanding the size of the dataset and optimizing the model structure in the future.

This study not only provides an efficient question and answer tool for the field of bridge design, but also provides a valuable reference and methodology for the construction of question and answer system in other professional fields.